\pdfoutput=1
\documentclass[11pt]{article}

\usepackage[]{acl}

\usepackage{times}
\usepackage{latexsym}

\usepackage[T1]{fontenc}
\usepackage[utf8]{inputenc}

\usepackage{microtype}

\usepackage{microtype}

\usepackage{verbatim}
\usepackage{booktabs, multirow} % for borders and merged ranges
\usepackage{subfig}
\usepackage{soul}% for underlines
\usepackage{graphicx} %插入图片的宏包
\usepackage{changepage,threeparttable} % for wide tables

\usepackage[T1]{fontenc}
\usepackage[utf8]{inputenc}
\usepackage{makecell}
\usepackage{threeparttable}
\usepackage{amsmath}
\usepackage{microtype}
\usepackage{enumerate}
\newenvironment{itemize*}%
 {\leftmargini=10pt\begin{itemize}%
  \setlength{\itemsep}{0pt}%
  \setlength{\parskip}{0pt}%
  }%
 {\end{itemize}}
\newenvironment{enumerate*}%
 {\begin{enumerate}%
  \setlength{\itemsep}{0pt}%
  \setlength{\parskip}{0pt}}%
 {\end{enumerate}}

\definecolor{myblue}{rgb}{0.9, 0.1, 0.94}
\definecolor{mygreen}{rgb}{0.64, 0.76, 0.68}
\definecolor{myyellow}{rgb}{0.88, 0.54, 0.35}
\definecolor{mygreen}{rgb}{0.68, 0.85, 0.9}
\definecolor{cyan}{rgb}{122, 87, 138}

\definecolor{myorange}{rgb}{1.0, 0.49, 0.0}

\title{Are All the Datasets in Benchmark Necessary? \\A Pilot Study of Dataset Evaluation for Text Classification}

\author{
    Yang Xiao$^{1}$, \;\; 
   Jinlan Fu$^{2 }$\thanks{\ \ Corresponding authors}, \;\; 
   See-Kiong Ng$^{2}$, \;\; 
   Pengfei Liu$^{3}$ 
   \\ 
   $^{1}$Fudan University,
 $^{2}$National University of Singapore, 
  $^{3}$Carnegie Mellon University \\ 
  \small{
\texttt{yangxiaocq12@gmail.com}, \;\; 
  \texttt{\{jinlan,seekiong\}@nus.edu.sg}, \;\; 
  \texttt{pliu3@cs.cmu.edu} 
  }
  }

\begin{document}
\maketitle
\begin{abstract}
In this paper, we ask the research question of whether all the datasets in the benchmark are necessary. We approach this by first characterizing the distinguishability of datasets when comparing different systems. Experiments on $9$ datasets and $36$ systems show that several existing benchmark datasets contribute little to discriminating top-scoring systems, while those less used datasets exhibit impressive discriminative power. We further, taking the text classification task as a case study, investigate the possibility of predicting dataset discrimination based on its properties (e.g., average sentence length).
Our preliminary experiments promisingly show that given a sufficient number of training experimental records, a meaningful predictor can be learned to estimate dataset discrimination over unseen datasets.
We released all datasets with features explored in this work on DataLab. \footnote{ \url{https://datalab.nlpedia.ai}}

\end{abstract}

\section{Introduction}

In natural language processing (NLP) tasks, there are often datasets that we use as benchmarks against which to evaluate machine learning models, either explicitly defined such as GLUE~\cite{wang-etal-2018-glue} and XTREME~\cite{DBLP:journals/corr/abs-2003-11080} 
or implicitly bound to the task (e.g., DPedia~\cite{DBLP:journals/corr/ZhangZL15} has become a default dataset for evaluating of text classification systems).
Given this mission, one important feature of a good benchmark dataset is the ability to statistically differentiate diverse systems \cite{DBLP:journals/corr/abs-2104-02145}.
With large pre-trained models consistently improving state-of-the-art performance on NLP tasks \cite{DBLP:journals/corr/abs-1810-04805,Lewis2019BARTDS}, the performances of many of them have reached a plateau \cite{zhong-etal-2020-extractive,DBLP:conf/aaai/FuLZ20}. In other words, it is challenging to discriminate a better model using existing datasets \cite{wang2019superglue}.
In this context, we ask the question: \textit{are all benchmark's datasets necessary?}
We use the text classification task as a case study and try to answer the following two sub-questions:

\begin{figure}[t]
    \centering
    {
    \includegraphics[width=0.8\linewidth]{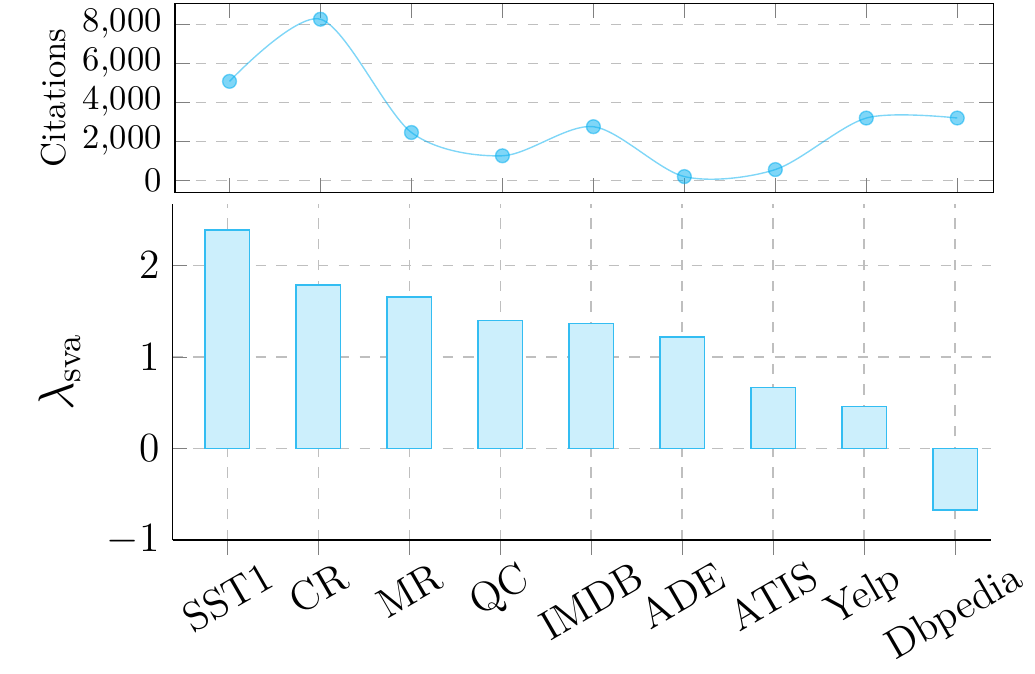} 
    }  
    \caption{Illustrate different datasets' distinguishing ability w.r.t top-scoring systems characterized by our measure $\mathrm{log}(\lambda_{\text{sva}})$ on text classification and their corresponding citations.}
    \label{fig:sva}
\end{figure}

\textbf{RQ1:} 
\textit{How can we quantify the distinguishing ability of benchmark datasets?}
To answer this question, we first design measures with varying calculation difficulties (\S\ref{sec:q1}) to judge datasets' discrimination ability based on top-scoring systems' performances.
By exploring correlations among different measures, we then evaluate how reliable a dataset's discrimination is when discrimination is calculated solely based on overall results that top-scoring systems have achieved and generalize this measure to other NLP tasks.
Fig.~\ref{fig:sva} illustrates how different text classification datasets are ranked (the bottom one) based on measures devised in this work (a smaller value suggests lower discrimination) and the corresponding citations of these datasets  (the upper one). One can observe that: 
(i) The highly-cited dataset \texttt{DBpedia} \cite{DBLP:journals/corr/ZhangZL15} (more than 3,000 times since 2015) shows the worst discriminative power.
(ii) By contrast, dataset like \texttt{ADE} \cite{GURULINGAPPA2012885} (less than 200 times since 2012) does  better in distinguishing top-scoring systems, suggesting that some of the relatively neglected datasets are actually valuable in distinguishing models. 
This phenomenon shows the significance of quantifying the discriminative ability of datasets: it can not only help us to \textbf{eliminate} those with lower discrimination from \textit{commonly-used datasets} (e.g., \texttt{DBpedia}), but also help us to \textbf{recognize} the missing pearl in \textit{seldom used} datasets (e.g., \texttt{ADE} and \texttt{ATIS} \cite{hemphill1990atis}).

\textbf{RQ2:} \textit{Can we try to predict the discriminative power of the dataset?}
Given a dataset, we investigate if we can judge its ability to distinguish models based on its characteristics (e.g., average sentence length), which is motivated by the scenario where a new dataset has just been constructed without sufficient top-scoring systems to calculate discrimination defined in RQ1.
To answer this question, inspired by recent literature on performance prediction \cite{domhan2015speeding,turchi2008learning,birch-etal-2008-predicting,xia-etal-2020-predicting,ye-etal-2021-towards}, we conceptualize this problem as a \textit{discrimination regression task}.
We define 11  diverse features to characterize a text classification dataset and regress its discrimination scores using different parameterized models.
Preliminary experiments (\S\ref{sec:regression-results}) indicate that a meaningful regressor 
can be learned to estimate the discrimination of unseen datasets without actual training using top-scoring systems.

We brief \textbf{takeaways} in this work based on our observations:

(1) Not all datasets in \textit{benchmark} are necessary in terms of model selection\footnote{Caveat: Annotated datasets are always valuable, because the supervision signals provided there can not only help us directly train a system for specific use case, but also provide good supervised transfer for related tasks \cite{sanh2021multitask}.}:  empirical results show that following datasets struggle at discriminating current top-scoring systems: \texttt{STS-B} and \texttt{SST-2} from GLUE \cite{wang-etal-2018-glue}; \texttt{BUCC} and \texttt{PAWX-X} from XTREME, which is consistent with the concurrent work \cite{DBLP:journals/corr/abs-2104-07412} (\S\ref{sec:other-benchmark}).

(2) In regard to single-task benchmark datasets, for Chinese Word Segmentation task, there are multiple datasets (\texttt{MSR, CityU, CTB}) \cite{tseng2005conditional,jin2008fourth} that exhibit much worse discriminative ability, suggesting that: future works on this task are encouraged to either (i) adopt other datasets to evaluate their systems or (ii) at least make significant test \footnote{We randomly select 10 recently published papers (from ACL/EMNLP) that utilized these datasets and found only 2 of them perform significant test.} if using these datasets. Similar observations happen in the dataset \texttt{CoNLL-2003} \cite{sang2003introduction} from Named Entity Recognition task and \texttt{MultiNLI} \cite{williams2017broad} from natural language inference task (\S\ref{sec:other-benchmark}).

(3) Some seldom used datasets such as \texttt{ADE} from text classification are actually better at distinguishing top-performing systems, which highlights an interesting and necessary future direction: \textit{how to identify infrequently-used but valuable (better discrimination) datasets for NLP tasks, especially in the age of dataset's proliferation?\footnote{\url{https://paperswithcode.com/datasets}}} (\S\ref{sec:tc_discrimination})

(4) Quantifying a dataset's discrimination (w.r.t top-scoring systems) by calculating the statistical measures (defined in \S\ref{sec:measure}) from leaderboard's results is a straightforward and effective way. But for those datasets without rich leaderboard results,\footnote{
The measure can keeps updated as the top-scoring systems of the leaderboard evolves, which can broaden its practical applicability} predicting the discrimination based on datasets' characteristics would be an promising direction (\S\ref{sec:popular_benchmark}).

Our \textbf{contributions} can be summarized as:

(1) We try to quantify the discrimination ability  for datasets by designing two variance-based measures.
(2) We systematically investigate $4$ text classification models on $9$ datasets, providing the newest baseline performance for those seldom used datasets. All datasets and their features are released on DataLab~\cite{xiao2022datalab}.
(3) We study several popular NLP benchmarks, including GLUE, XTREME, NLI, and so on. Some valuable suggestions and observations will make research easier.

\section{Related Work}

\paragraph{Benchmarks for NLP}
In order to conveniently keep themselves updated with the research progress, researchers recently are actively  building evaluation benchmarks for diverse tasks so that they could make a comprehensive comparison of systems, and use a leaderboard to record the evolving process of the systems of different NLP tasks,
such as SQuAD~\cite{rajpurkar-etal-2016-squad}, GLUE~\cite{wang-etal-2018-glue}, XTREME~\cite{DBLP:journals/corr/abs-2003-11080}, GEM~\cite{gehrmann2021gem} and GENIE~\cite{khashabi2021genie}.
Despite their utility, more recently, \citet{DBLP:journals/corr/abs-2104-02145} highlight that
unreliable and biased systems score so highly on standard benchmarks that there is little room for
researchers who develop better systems to
demonstrate their improvements. In this paper, we make a pilot study on meta-evaluating benchmark evaluation datasets and quantitatively characterize their discrimination in different top-scoring systems. 

\paragraph{Performance Prediction}
Performance prediction is the task of estimating a system's performance without the actual training process.
With the recent booming of the number of machine learning models \cite{Goodfellow-et-al-2016} and datasets, the technique of performance prediction become rather important when applied to different scenarios ranging from early stopping training iteration \cite{kolachina-etal-2012-prediction}, architecture searching \cite{domhan2015speeding}, and attribution analysis \cite{birch-etal-2008-predicting,turchi2008learning}.
In this work, we aim to calculate a dataset's discrimination without actual training top-scoring systems on it, which can be formulated as a performance prediction problem.

\section{Preliminaries}

\subsection{Task and Dataset}

Text classification aims to assign a label defined beforehand to a given input document. In the experiment, we choose nine datasets, and their statistics can be found in the Appendix \ref{sec:app-statistics}.
\begin{itemize*}
    \item \textbf{IMDB} \cite{maas-etal-2011-learning} consists of movie reviews with binary classes.
    \item \textbf{Yelp} \cite{DBLP:journals/corr/ZhangZL15} is a part of the Yelp Dataset Challenge 2015 data.
    \item \textbf{CR} \cite{10.1145/1014052.1014073} is a product review dataset with binary classes.
    \item \textbf{MR} \cite{DBLP:journals/corr/abs-cs-0506075} is a movie review dataset collected from Rotten Tomatoes. 
    \item \textbf{SST1}
    \cite{socher-etal-2013-recursive} is collected from HTML files of Rotten Tomatoes reviews with fully labeled parse trees. 
    \item \textbf{DBpedia14} \cite{DBLP:journals/corr/ZhangZL15} is a dataset for ontology classification collected from DBpedia. 
    \item \textbf{ATIS} \cite{hemphill1990atis} is an intent detection dataset that contains audio recordings of flight reservations. 
    \item \textbf{QC} \cite{li-roth-2002-learning} is a question classification dataset.
    \item \textbf{ADE} \cite{GURULINGAPPA2012885} is a subset of ``Adverse Drug Reaction Data''. 
\end{itemize*}

\subsection{Model}
\label{sec:model}

We re-implement $4$ top-scoring systems with typical neural architectures for each dataset.~\footnote{We mainly focus on neural network-based models, since most top-scoring systems in the leaderboard are based on deep learning. } The brief introduction of the four models is as follows. 

\begin{itemize*}
\item \textbf{LSTM} \cite{hochreiter1997long} is a widely used sentence encoder. Here, we adopt the bidirectional LSTM. 
\item \textbf{LSTMAtt} is proposed by \citet{DBLP:journals/corr/LinFSYXZB17} that designed the self-attention mechanism to extract different aspects of features for a sentence.
\item \textbf{BERT} \cite{DBLP:journals/corr/abs-1810-04805} was utilized to fine-tuning on our text classification datasets.
\item \textbf{CNN} is a CNN-based text classification model \cite{DBLP:journals/corr/Kim14f} was expolred in our work.
\end{itemize*}

Except for BERT, the other three models (e.g. \texttt{LSTM})  are initialized by GloVe \cite{DBLP:conf/emnlp/PenningtonSM14} or Word2Vec \cite{DBLP:conf/nips/MikolovSCCD13} pre-trained word embeddings. When the performance on the dev set doesn't improve within 20 epochs, the training will be stopped, and the best performing model will be kept. 
More detailed model parameter settings can be found in the Appendix \ref{sec:app-param}.

\section{How to Characterize Discrimination?} \label{sec:q1}
To achieve this goal, we design measures based on the performance of different models for a dataset.

\subsection{Measures} \label{sec:measure}

We design several measures to judge  dataset's distinguishing ability based on the performances that top-performing systems have achieved on it.\footnote{A dataset's discrimination is defined w.r.t top-scoring models from a leaderboard, keeping itself updated with systems' evolution.}
Specifically, given a dataset $D$ together with $k$ top-scoring model \textit{performance list} $\mathbf{v} = [v_1, \cdots, v_k]$, we define the following measures.

\subsubsection{Performance Variance} We use the standard deviation  to quantify the degree of variation or dispersion of a set of performance values. A larger value of $\lambda_{\text{var}}$ suggests that the discrimination of the given dataset is more significant.
$\lambda_{\text{var}}$ can be defined as: 
\begin{equation}
\lambda_{\text{var}} = \text{Std}(\mathbf{v}), 
\end{equation}
where $\text{Std}(\cdot)$ is the function to compute the standard deviation. Assume that the performance list ($k=3$) on dataset $D$ is $\mathbf{v} = [88,92,93]$, we can get  $\lambda_{\text{var}}=2.65$.    

\subsubsection{Scaled Performance Variance} \label{sec:measure}
For the above measure, it can only reflect the variances of the performance of different models, without considering whether the model's performance is close to the upper limit (e.g., 100\% accuracy) on a given data set.
To address this problem, we defined a modified variance by scaling $\lambda_{\text{var}}$ with the difference between the upper limit performance ${u}$ and average performance $\mathrm{Avg}(\mathbf{v})$ of $\mathbf{v}$.

\begin{equation}
    \lambda_{\text{sva}} = \lambda_{\text{var}}  ({u} -\mathrm{Avg}(\mathbf{v})).
\end{equation}
In practice, ${u}$ can be defined flexibly based on tasks' metrics. For example, in text classification task, ${u}$ could be $100\%$ (w.r.t F1 or accuracy), while in summarization task, ${u}$ could be the results of oracle sentences (w.r.t ROUGE). 
Intuitively, given a performance list on text classification dataset: $\mathbf{v} = [88,92,93]$, we can obtain the  $\lambda_{\text{sva}}=23.81$.

\begin{table*}[!ht]\centering
\small
\begin{tabular}{lcccccccc}\toprule
Method&BERT &LSTMAttr &LSTM &CNN  &$\lambda_{\text{hit}}$ &$\lambda_{\text{var}}$ &$\lambda_{\text{sva}}$  \\\midrule
SST1 &54.12 &43.80 &47.60 &44.80 &0.88 &4.65 &243.56 \\
CR &91.75 &83.25 &82.50 &84.25 &0.91 &4.27 &62.17 \\
MR &85.55 &79.92 &79.80 &82.00 &0.86 &2.69 &48.83 \\
QC &97.19 &90.36 &89.96 &92.17 &0.92 &3.32 &25.18 \\
IMDB &93.34 &89.45 &89.65 &87.81 &0.87 &2.33 &23.18 \\
ADE &93.48 &92.90 &92.65 &89.54 &0.78 &1.77 &13.90 \\
ATIS &97.64 &97.42 &97.31 &94.62 &0.78 &1.42 &4.63 \\
Yelp &97.52 &96.60 &96.60 &95.46 &0.81 &0.84 &2.91 \\
DPedia &99.27 &99.01 &99.05 &98.75 &0.68 &0.22 &0.21 \\
\midrule
Spearman & & & & & &0.83 &0.73 \\
\bottomrule
\end{tabular}
\caption{Illustration the $4$  models' performance and discrimination degree (characterized by $\lambda_{\text{hit}}$, $\lambda_{\text{var}}$, and $\lambda_{\text{sva}}$) on $9$ text classification datasets. The two correlation coefficients pass the significance test ($p<0.05$ ). 
$\lambda_{\text{var}}$ and $\lambda_{\text{sva}}$ measures are designed based on performance variance. 
}
\label{tab:tc_hit}
\end{table*}

\subsubsection{Hit Rate}
\label{sec:hit}
The previous two measures quantify dataset's discriminative ability w.r.t $k$ top-performing systems in an \textit{indirect} way (i.g, solely based on the overall results of different models). However, sometimes, small variance does not necessarily mean that the dataset fail to distinguish models, as long as the difference between models is statistically significant.
To overcome this problem, we borrow the idea of bootstrap-based significant test \cite{koehn-2004-statistical} and define the measure \textit{hit rate}, which quantify the degree to which a given dataset could successfully differentiate $k$ top-scoring systems.

Specifically, we take all $\tbinom{k}{2}$ pairs of systems ($m_i$ and $m_j$) and compare their performances on a subset of test samples $D_t$ that is generated using paired bootstrap re-sampling.
Let $v_i(D) > v_j(D)$ be the performance of $m_1$ and $m_2$ on the full test set, we define $P(m_i, m_j)$ as the frequency of $v_i(D_t) > v_j(D_t)$ over all $T$ times of re-sampling ($t = 1, \cdots, T$).~\footnote{For example, given a test set with 1000 samples, we sample 80\% subset from it and repeat this process T times.}
Then we have
\begin{equation}
    \lambda_{\text{hit}} = \frac{1}{\tbinom{k}{2}}\sum P(m_i, m_j)
\end{equation}

\paragraph{Metric Comparison}
The first two metrics, performance variance  and scaled performance variance, are relative easily to obtain since they only require holistic performances of different top-scoring models on a given dataset, which can be conveniently collected from existing leaderboards. By contrast, although the metric \textit{hit rate} can directly reflect dataset's ability in discriminating diverse systems, its calculation not only require more fine-grained information of system prediction but also complicated bootstrap re-sampling process.

\subsection{Exp-I: Exploring Correlation Between Variance and Hit Rate}

The goal of this experiment is to investigate the reliability of the variance-based discrimination measures (e.g., $\lambda_{\text{sva}}$), which are easier to obtain, by calculating its correlation with significant test-based measure  $\lambda_{\text{hit}}$, which is costly to get.
Since the implementation of $\lambda_{\text{hit}}$ relies on the  bootstrap-based significant test, we choose text classification as the tested and re-implement $4$ classification models (defined in Sec.~\ref{sec:model}) on $9$ datasets. The performance and the distinction degree on the $9$ text classification dataset are shown in Tab.~\ref{tab:tc_hit}.  $\lambda_{\text{var}}$ and $\lambda_{\text{sva}}$ measures are designed based on performance variance, 
even if BERT always achieves the best performance on the same dataset, it will not affect the observed results from our experiments. 

\paragraph{Correlation measure} 
Here, we adopt the Spearman rank correlation coefficient \cite{zar1972significance} to describe the  correlation between our variance-based measures and the hit rate measure $\lambda_{\text{hit}}$.
\begin{equation}
    S_{\lambda} = \text{Spearman}(q, \lambda_{\text{hit}}),
\end{equation}
where the $q$ can be $\lambda_{\text{var}}$ or $\lambda_{\text{sva}}$.

\paragraph{Result}
\label{sec:tc_discrimination}

\noindent
(1) $\lambda_{\text{var}}$  and $\lambda_{\text{sva}}$ are strong correlative ($S_{\lambda}$>0.6) with $\lambda_{\text{hit}}$ respectively, which suggests that 
variance-based metrics could be a considerably reliable alternatives of significant test-based metric. 

\noindent
(2) $\text{Spearman}(\lambda_{\text{var}}, \lambda_{\text{hit}}) > \text{Spearman}(\lambda_{\text{sva}},
\lambda_{\text{hit}})$, which indicate that comparing with $\lambda_{\text{sva}}$, 
dataset discrimination characterized by $\lambda_{\text{var}}$ is more acceptable for $\lambda_{\text{hit}}$. 
The reason can be attributed to that the designing of the measure $\lambda_{\text{hit}}$ does not consider the upper limit of the model's performance.

\noindent
(3) \texttt{DPdedia} and \texttt{Yelp} are commonly used text classification datasets, while they have the worst ability to discriminate the top-scoring models since they get the lowest value of $\lambda_{\text{var}}$ and $\lambda_{\text{sva}}$. By contrast, these two seldom used datasets \texttt{ADE} and \texttt{ATIS} show the better discriminative ability.

\subsection{Exp-II: Evaluation of Other Benchmarks}

\subsubsection{Popular Benchmark Datasets}
\label{sec:popular_benchmark}
We also investigate how benchmark datasets from other NLP task perform using two devised measures.
Specifically, we collected three single-task and two multitask benchmarks.
For the single-task benchmarks, we collect the top-performing models in a specific period for each dataset, provided by Paperswithcode.~\footnote{\url{https://paperswithcode.com/}}
For the multitask benchmarks, here, the GLUE~\footnote{\url{https://gluebenchmark.com/}}  and XTREME~\footnote{\url{https://sites.research.google/xtreme}} are considered in this work.
Since Paperswithcode provided $5$ models for each dataset in most case, for fairness and uniformity, we keep top-5 models for both single-task and multitask benchmark datasets.

\begin{figure*}
    \centering 
     \subfloat[{GLUE}]{
    \includegraphics[width=0.44\linewidth]{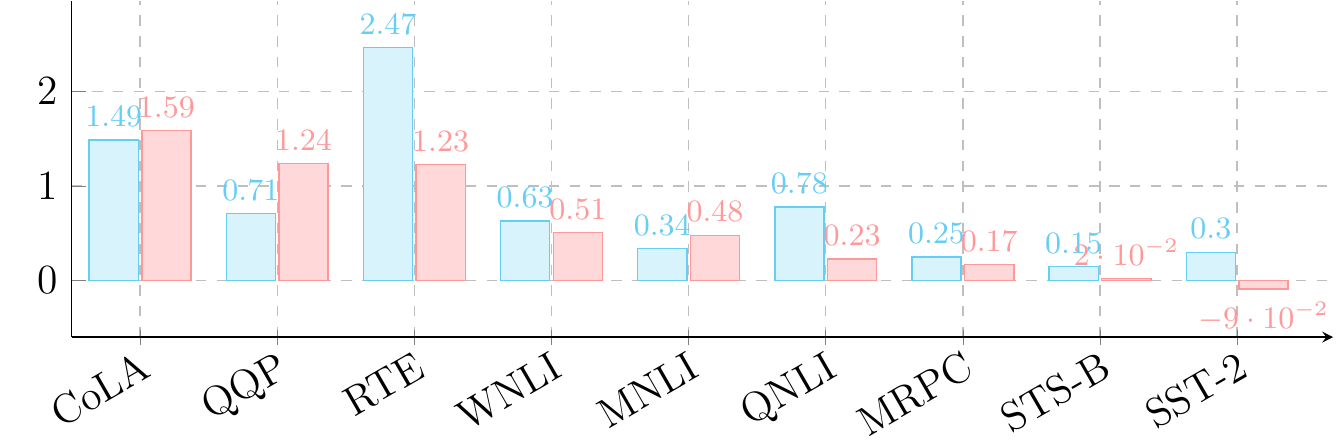}
    } 
    \subfloat[{XTREME}]{
    \includegraphics[width=0.44\linewidth]{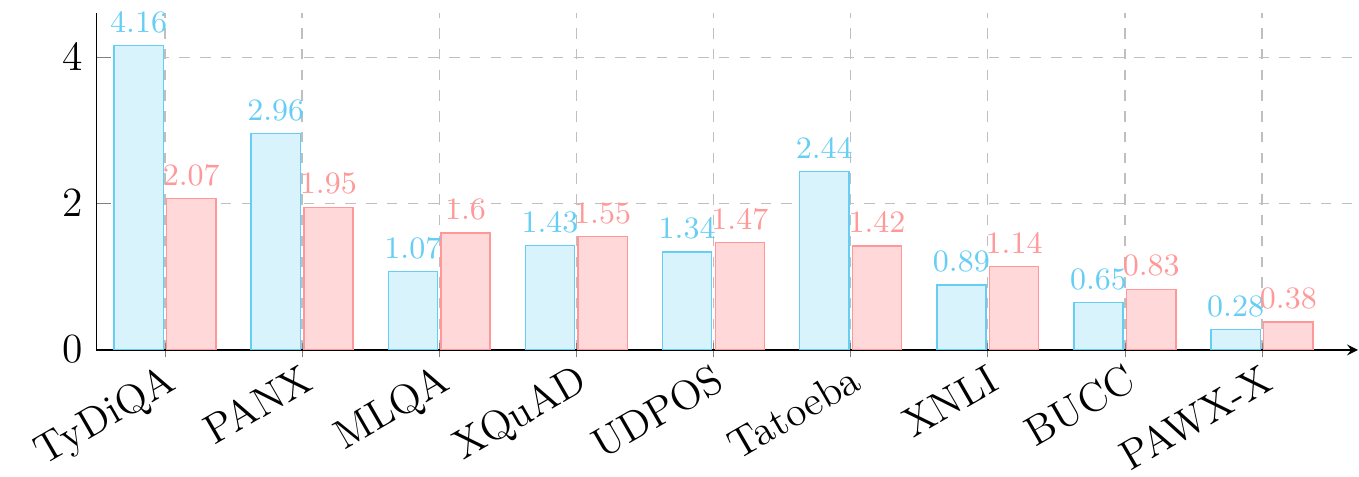}
    }  \\
    \subfloat[{CWS}]{
    \includegraphics[width=0.37\linewidth]{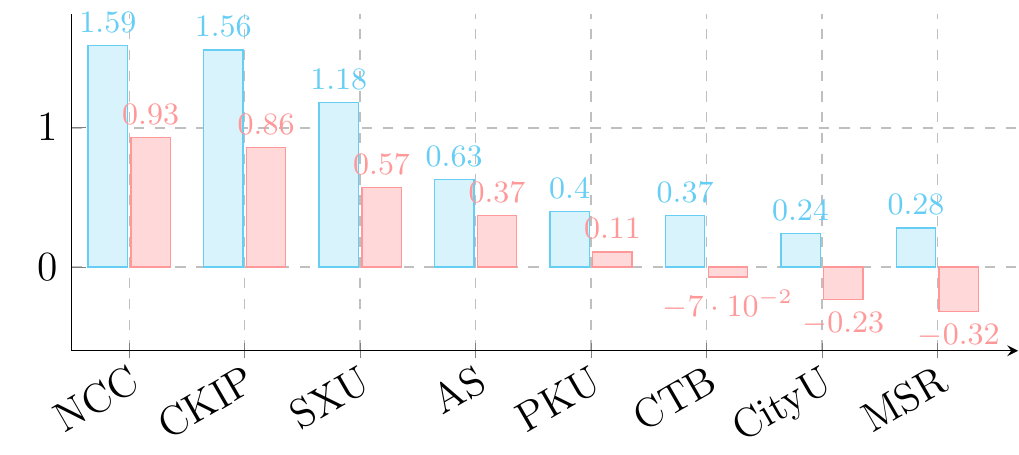}
    }  
    \subfloat[{NER}]{
    \includegraphics[width=0.3\linewidth]{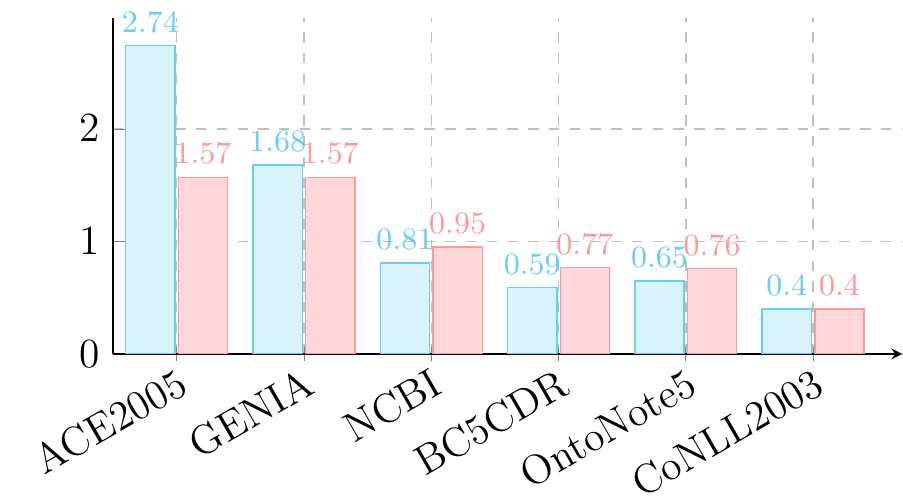} 
    } 
    \subfloat[{NLI}]{
    \includegraphics[width=0.21\linewidth]{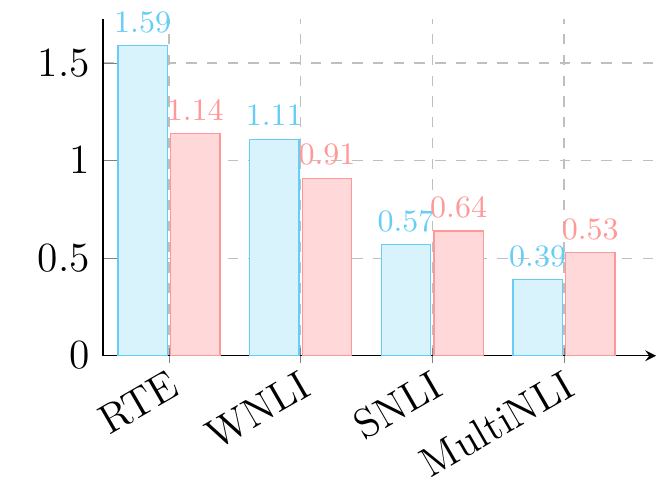}
    } 
    \caption{The dataset discrimination characterized by $\mathrm{log}(\lambda_{\text{var}})$ (the logarithm for better visualization) (blue) and $\mathrm{log}(\lambda_{\text{sva}})$ (pink) on five popular NLP benchmarks.}
    \label{fig:histogram-quality}
\end{figure*}

\noindent \textbf{Named Entity Recognition (NER)} aims to identify named entities of an input text, for which we choose $5$ top-scoring systems on $6$ datasets and collect  results from Paperswithcode.

\noindent \textbf{Chinese Word Segmentation (CWS)}
aims to detect the boundaries of Chinese words in a sentence. We select $5$ top-scoring systems on  $8$ datasets and collect results from Paperswithcode.

\noindent \textbf{Natural Language Inference (NLI)}
targets at predicting whether a premise sentence can infer the hypothesis sentence. We select $5$ top-performing models on $4$ datasets from Paperswithcode.

\noindent \textbf{GLUE} \cite{wang-etal-2018-glue} covers $9$ sentence- or sentence-pair tasks with different dataset sizes, text genres, and degrees of difficulty. 
Fig.~\ref{fig:histogram-quality}-(a) shows the tasks/datasets that are considered in GLUE.

\noindent \textbf{XTREME} \cite{DBLP:journals/corr/abs-2003-11080}
is the first benchmark that evaluates models across a  wide variety of languages and tasks. 
The tasks/datasets that are covered by XTREME are shown in Fig. \ref{fig:histogram-quality}-(b).

\subsubsection{Results and Analysis}
\label{sec:other-benchmark}

Fig.~\ref{fig:histogram-quality}
shows the results of dataset quality measure by $\lambda_{\text{var}}$ and $\lambda_{\text{sva}}$. We detail several main observations:

\begin{itemize*}
    \item $\lambda_{\text{var}}$ and $\lambda_{\text{sva}}$  have consistent evaluation results for both single-task (CWS, NER, NLI) and multitask (GLUE, XTREME) benchmarks. 
    \item For the XTREME benchmark, \texttt{BUCC} and \texttt{PAWSX} have lowest $\lambda_{\text{var}}$ and $\lambda_{\text{sva}}$, which suggest that they are hardly to discriminate the top-performing systems. Moreover, these two data sets will be removed from the new version of the XTREME leaderboard called XTREME-R \cite{DBLP:journals/corr/abs-2104-07412}. This consistent observation also shows the effectiveness of our measure.
    \item For GLUE benchmark, \texttt{CoLA}, \texttt{QQP}, and \texttt{RTE} have the excellent ability to distinguish different top-scoring models (with higher $\lambda_{\text{var}}$ and $\lambda_{\text{sva}}$), while the \texttt{SST-2} and \texttt{STS-B} perform worse.
    \item For CWS benchmarks, there is a larger gap between the value of $\lambda_{\text{var}}$ and $\lambda_{\text{sva}}$, which indicate that the performance of top-scoring models considered are close to 100\%. 
    Furthermore, \texttt{MSR}, \texttt{CityU} and \texttt{CTB} are not suitable as benchmarks since they have poor discrimination ability with $\lambda_{\text{sva}} <0$. So as \texttt{MultiNLI} for NLI task.
    \item \texttt{CoNLL 2003} is a widely used NER dataset, but it is the lowest quality dataset under our dataset quality measure. The reason can be attributed to contain much annotation errors \cite{DBLP:conf/aaai/FuLZ20} in the \texttt{CoNLL 2003} dataset, which makes its performance reach the bottleneck.  
    
\end{itemize*}

\section{Can we Predict Discrimination?}
Although metrics $\lambda_{\text{var}}$, $\lambda_{\text{sva}}$ ease the burden for us to calculate the datasets' discrimination, one major limitation is: given a new dataset without results from leaderboards, we need to train multiple top-scoring systems and calculate corresponding results on it, which is computationally expensive.
To alleviate this problem, in this section, we focus on text classification task and investigate the possibility of estimating datasets' discrimination solely based on their characteristics without actual training systems on them.

\subsection{Task Formulation}

\subsubsection{Regression-based Task Formulation}
We formulate it as a performance prediction problem \cite{birch-etal-2008-predicting,xia-etal-2020-predicting,ye-etal-2021-towards}.
Formally, we refer to $\mathcal{M}$, $D^{tr}$ , $D^{te}$, $\mathcal{S}$ as the machine learning system, training data, test data and training strategy respectively. 
The goal of performance prediction is to estimate actual performance $y$ without actual training by using features of $\mathcal{M}$, $\mathcal{D}^{tr}$, $\mathcal{D}^{te}$, and $\mathcal{S}$.

\begin{equation}
    \hat{y} = \hat{f}(\Phi_{\mathcal{M}}, \Phi_{\mathcal{D}^{tr}}, \Phi_{\mathcal{D}^{te}}, \Phi_{\mathcal{S}};\hat{\Theta})
\end{equation}

where $\hat{y}$ denotes estimated prediction and $\Phi(\cdot)$ is a feature extractor. Following \citealt{xia-etal-2020-predicting}, we only use the features of the datasets as variables and adapt it to our discriminative prediction scenario, we can obtain:

\begin{equation}
    \hat{\lambda} = \hat{f}(\Phi_{\mathcal{D}^{tr}}, \Phi_{\mathcal{D}^{te}};\hat{\Theta}),
    \label{eq:regressor}
\end{equation}
where $\hat{\lambda}$ denotes predicted variance defined in \S\ref{sec:measure} such as $\lambda_{var}$ or $\lambda_{sva}$.

\subsubsection{Ranking-based Task Formulation}
\label{sec:ranking_based task formulation}
Instead of only regressing one dataset's quality, we also care about the quality ranking of different datasets w.r.t discriminating systems in a task. Therefore, we also formulate it as a listwise LTR(learning to rank) task where a model takes individual lists as instances, to predict the rank of element among the list \cite{liu2011learning}.
Given a set of $n$ datasets $d = \{d_1,d_2,\cdots,d_n \}$ ($d \in D=\{D^{tr}, D^{te}\}$), different $d$ construct the dataset of LTR task, the target of the ranker is to predict the dataset quality ranking for each dataset in $d$ according to the datasets' features. The estimated rankings $\overline{\mathbf{\lambda}} = \{\lambda_1, \lambda_2, \cdots, \lambda_n\} \in [1,n]$ for set $d$ can be defined as:

\begin{equation}
\label{eq:ranking}
    \overline{\mathbf{\lambda}} = \overline{f}( \Phi_{(d)};\overline{\Theta}),
\end{equation}
where $\Phi(\cdot)$ is the dataset feature extractor, $\overline{f}$ is the ranking model.  $\overline{\mathbf{\lambda}} \in [1,n]$ is the estimated rankings of the variance ( $\lambda_{var}$ or $\lambda_{sva}$) for datasets in set $d$.

\subsection{Characterization of Datasets}
\label{sec:data-feature}
In this section, we will introduce three aspects that characterize datasets: Inherent Feature,  Lexical Feature, and Semantic Feature. Due to space limitations, we move a more detailed feature introduction to the Appendix \ref{sec:app-feat}.

\subsubsection{Inherent Feature}

\noindent\textbf{Average length ($\phi_{\mathrm{len}}$):} The average sentence length on a dataset, where the number of tokens on a sentence is considered as the sentence length. 
\noindent\textbf{Label number ($\phi_{\mathrm{lab}}$):} The number of labeled classes in a dataset.
\noindent\textbf{Label balance ($\phi_{\mathrm{bal}}$):} The label balance metric measures the variance between the ideal and the true label distribution.

\subsubsection{Lexical Feature}

\noindent\textbf{Basic English Words Ratio ($\phi_{\mathrm{basic}}$):}
The proportion of words belonging to the 1000 basic English~\footnote{\url{https://simple.wikipedia.org/wiki/Wikipedia:List_of_1000_basic_words}} words in the whole dataset. 
\noindent\textbf{Type-Token Ratio ($\phi_{\mathrm{ttr}}$):}
We measure the text lexical richness by the type-token ratio \citep{richards_1987} based on the lexical richness tool.~\footnote{\url{https://github.com/LSYS/lexicalrichness}}
\noindent\textbf{Language Mixedness Ratio ($\phi_{\mathrm{lmix}}$):}
To detect the ratio of other languages mixed in the text, we utilize the models proposed by \citet{joulin2016bag} for language identification from fastText~\citep{joulin2016fasttext} which can recognize 176 languages.
\noindent\textbf{Pointwise Mutual Information ($\phi_{\mathrm{pmi}}$):} PMI~\footnote{\url{https://en.wikipedia.org/wiki/Pointwise_mutual_information}} is a measurement to calculate the correlation between variables.

\subsubsection{Semantic Feature}

\noindent\textbf{Perplexity ($\phi_{\mathrm{ppl}}$):}
We calculate the perplexity~\footnote{\url{https://en.wikipedia.org/wiki/Perplexity}} based on GPT2 \citep{radford2019language} to evaluate the quality of the text.
\noindent\textbf{Grammar Errors Ratio ($\phi_{\mathrm{gerr}}$):} We adopt the detection tool~\footnote{\url{https://github.com/jxmorris12/language_tool_python}} to recognize words with grammatical errors, and then calculate the ratio of grammatical errors. 
\noindent\textbf{Flesch Reading Ease~\footnote{\url{https://en.wikipedia.org/wiki/Flesch\%E2\%80\%93Kincaid_readability_tests}} ($\phi_{\mathrm{fre}}$):} 
To describe the readability of a text, we introduce the  $\phi_{\mathrm{fre}}$ achieving by textstat.~\footnote{\url{https://github.com/shivam5992/textstat}}

For feature $\phi_{\mathrm{len}}$, $\phi_{\mathrm{ttr}}$,$\phi_{\mathrm{lmix}}$,  $\phi_{\mathrm{gerr}}$,  $\phi_{\mathrm{pmi}}$, $\phi_{\mathrm{fre}}$, and $\phi_{\mathrm{r_{fre}}}$, we individually compute $\phi()$ on the training, test set, as well as their interaction.
Take average length ($\phi_{\mathrm{len}}$) as an example, we compute the average length on training set $\phi_{\mathrm{tr,len}}$, test set $\phi_{\mathrm{te,len}}$, and their interaction $((\phi_{\mathrm{tr,len}}-\phi_{\mathrm{te,len}})/\phi_{\mathrm{tr,len}})^2$.

\subsection{Parameterized Models}
The dataset discrimination prediction (ranking) model takes a series of dataset features as the input and then predicts discrimination(rank) based on $\hat{f}(\cdot)$ ($\overline{f}(\cdot)$) defined in Eq.~\ref{eq:regressor} (Eq.~\ref{eq:ranking}).
We explore the effectiveness of four variations of regression methods and two ranking frameworks.

\noindent
\textbf{Regression Models}:
 \textbf{LightGBM} \cite{DBLP:conf/nips/KeMFWCMYL17} is a gradient boosting framework with faster training and better performance than XGBoost.
 \textbf{K-nearest Neighbor (KNN)} \cite{peterson2009k}
is a non-parametric model that makes the prediction by exploring the k neighbors. 
 \textbf{Support Vector Machine (SVM)} \cite{suykens1999least} uses kernel trick to solve both linear and non-linear problems. 
  \textbf{Decision Tree (DT)} \cite{quinlan1990probabilistic} is a tree-based algorithm that gives an understandable interpretation of predictions.

\noindent
\textbf{Ranking Frameworks:}
    \textbf{LightGBM} with Gradient Boosting Decision Tree \cite{friedman2001greedy} boosting strategy was selected as our ranking model.
      \textbf{XGBoost} \cite{2016} with gbtree\cite{Hastie2009} 
    boosting strategy was another ranking model.

\subsection{Experiments} \label{sec:regression-results}

\subsubsection{Data Construction}
To construct a collection with large amount of discriminative datasets, we randomly select three dataset features (e.g. average sentence length $\phi_{\text{len}}$) to divide the original dataset into several non-overlapping sub-datasets. As a result, we collect $987$ sub-datasets. Then, we train four text classification models (CNN, LSTM, LSTMAtt, BERT) on these sub-dastasets. 
Next, we calculate the dataset features $\phi$ (defined in Sec. \ref{sec:data-feature}) and dataset discrimination ability $\lambda_{\text{sva}}$ and  $\lambda_{\text{var}}$  on these sub-datasets. 

\noindent\textbf{Regression Task Settings}
$\phi$ and $\lambda_{\text{sva}}$ ($\lambda_{\text{var}}$) will be the input and target of the regression models, as defined by Eq. \ref{eq:regressor}. For the experiment setting, we randomly select 287 ($\phi$, $\lambda_{\text{sva}}$ ($\lambda_{\text{var}}$)) pairs as the test set and the rest as the training set (700). 
\noindent\textbf{Ranking Task Settings}
We construct datasets for ranking task from the dataset used in regression task. 
Here, we explored the value of $n$ (defined in \S\ref{sec:ranking_based task formulation}) to be $5$, $7$ and $9$ to randomly choose samples from $D^{tr}$ (or $D^{te}$) to construct the datasets for the ranking task, and kept $4200$, $600$, $1200$ samples for training, development and testing set respectively.

\subsubsection{Evaluation Metric}
\paragraph{Regression Task} We use  RMSE \cite{chai2014root} and Spearman rank correlation coefficient \cite{zar1972significance} to evaluate how well the regression model predicts the discriminative ability for datasets. The Spearman rank correlation coefficient is used for the correlation between the output of a regression model and the ground truth.

\paragraph{Ranking Task}
NDCG \cite{DBLP:conf/sigir/JarvelinK00} and MAP \cite{DBLP:conf/sigir/YueFRJ07} are the evaluation metric of our ranking task. For NDCG, it considers the rank of a set of discriminative abilities. In our setting, every dataset has its own real discriminative ability. Here, We transfer the predicted discriminative ability to the rank of the dataset in the NDCG metric, so we can use NDCG to evaluate the model’s predicted effect.
For MAP, it likes how NDCG works, but it considers a set of binary values.
Here, we set a threshold value of $\lambda_{\text{var}}=3$ ($\lambda_{\text{sva}}=28$) for $\lambda_{\text{var}}$ ($\lambda_{\text{sva}}$) to distinguish the dataset discrimination ability from good (relevant) to bad (irrelevant).

\begin{table}[htb]
  \centering \footnotesize
  \renewcommand\tabcolsep{3pt}
    \begin{tabular}{lcccccc}
    \toprule
    \multirow{3}[4]{*}{Method} & \multicolumn{2}{c}{RMSE} & \multicolumn{4}{c}{Spearman} \\
\cmidrule(lr){2-3}\cmidrule(lr){4-7}          & \multirow{2}[2]{*}{$\lambda_{\text{var}}$} & \multirow{2}[2]{*}{$\lambda_{\text{sva}}$} & \multicolumn{2}{c}{$\lambda_{\text{var}}$} & \multicolumn{2}{c}{$\lambda_{\text{sva}}$} \\
\cmidrule(lr){4-5}\cmidrule(lr){6-7} 
          &       &       & corr & p     & corr & p \\
    \midrule
    KNN   & 2.42  & 51.21  & 0.77  & 9.75E-40 & 0.87  & 1.62E-63 \\
    LightGBM & 1.53  & 32.74  & 0.72  & 2.23E-33 & 0.87  & 7.01E-61 \\
    DT & 1.73  & 43.33  & 0.64  & 9.25E-25 & 0.84  & 1.33E-53 \\
    SVM   & 2.83  & 62.44  & 0.68  & 1.14E-28 & 0.77  & 7.26E-40 \\
    \bottomrule
    \end{tabular}
     \caption{The performance of regressing dataset discrimination for the text classification. ``\textit{corr}'' denotes the ``\textit{correlation}''. 
     }
   \label{tab:regression_results}%
\end{table}

\begin{table}[!ht]\centering
\footnotesize
\begin{tabular}{lrrrrrr}\toprule
\multirow{2}{*}{Model} &\multirow{2}{*}{n} &\multicolumn{2}{c}{NDCG} &\multicolumn{2}{c}{MAP} \\\cmidrule(lr){3-4}\cmidrule(lr){5-6}
& &$\lambda_{\text{var}}$ &$\lambda_{\text{svar}}$ &$\lambda_{\text{var}}$ &$\lambda_{\text{svar}}$ \\\midrule
\multirow{3}{*}{LightGBM} &9 &98.20 &98.85 &97.50 &98.27 \\
&7 &97.76 &98.73 &97.01 &99.05 \\
&5 &96.73 &97.08 &96.56 &98.15 \\
\midrule
\multirow{3}{*}{XGBoost} &9 &96.66 &97.13 &92.91 &93.62 \\
&7 &96.74 &97.65 &94.77 &96.11 \\
&5 &95.93 &97.10 &95.49 &98.25 \\
\bottomrule
\end{tabular}
\caption{The performance of ranking dataset discrimination for the text classification task. \textit{n} is the number of datasets in $d$ 
defined in \S\ref{sec:ranking_based task formulation}}\label{tab: ranking_results}.
\end{table}

\subsubsection{Results and Analysis}
Tab.~\ref{tab:regression_results} and Tab.~\ref{tab: ranking_results} show the results of four regression models and two ranking models that characterize the dataset discrimination ability, respectively. We can observe that: Both the regression models and the ranking models can well describe the discrimination ability of different datasets.
For these four regression models, the prediction is highly correlated with the ground truth (with a correlation value larger than 0.6), passing the significance testing ($p<0.05$). This suggests that the dataset discrimination can be successfully predicted.
For these two ranking models, their performance on NDCG and MAP is greater than 95\%, which indicates that the discriminative ability of the data set can be easily ranked.

\begin{figure}[htb]
    \centering 
    \subfloat[{Inherent}]{
    \includegraphics[width=0.3\linewidth]{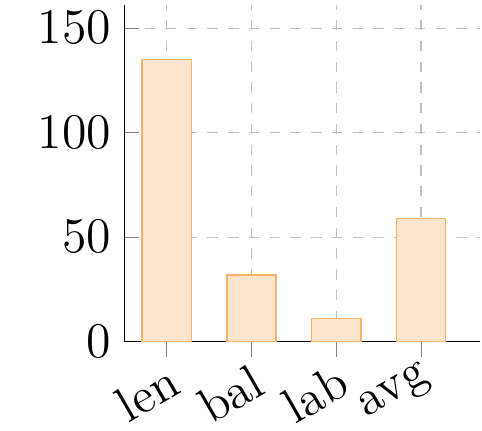}
    }  \hspace{-8pt}
    \subfloat[{Lexical}]{
    \includegraphics[width=0.34\linewidth]{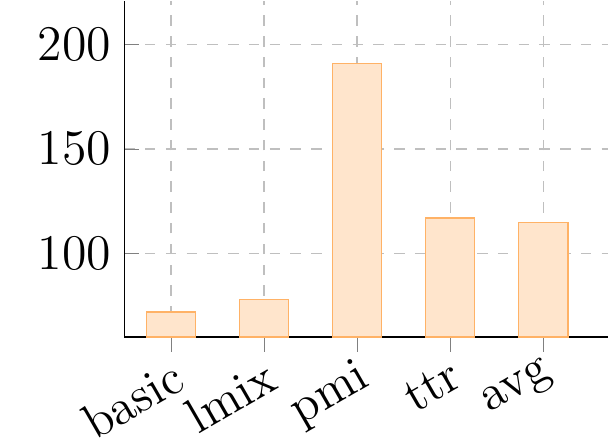} 
    } \hspace{-8pt}
    \subfloat[{Semantic}]{
    \includegraphics[width=0.3\linewidth]{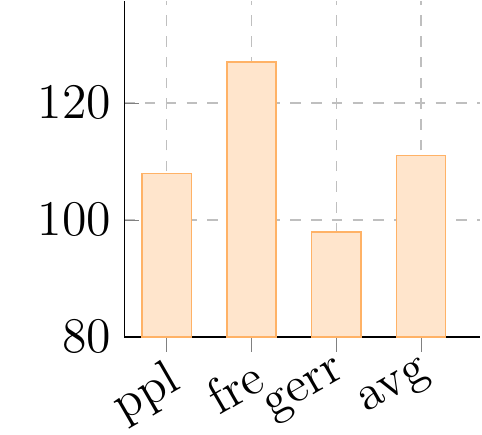}
    } 
    \caption{Feature importance for the text classification measured by LGBoost with the target of $\lambda_{\text{sva}}$.}
    \label{fig:feature_importance}
\end{figure}

\paragraph{Feature Importance Analysis}
Fig.~\ref{fig:feature_importance} illustrates the feature importance characterized by LightGBM. For a given feature, the number of times that is chosen as the splitting feature in the node of the decision trees is defined as its importance degree.
We observe that:
(1) The most influential features are $\phi_{\text{pmi}}$, $\phi_{\text{len}}$, and $\phi_{\text{fre}}$, which come from the lexical, inherent, and semantic features, respectively. This indicated that the LightGBM can extract features from different aspects to make predictions.
(2) In the perspective of feature groups, the semantic features are more influential than the inherent features and lexical features.

\section{Discussion \& Implications}
\paragraph{Discussion}
Given a leaderboard of a dataset, metrics explored in this paper can be easily used to calculate its discrimination, while some limitations still exist. We make some discussion below to encourage more explorations on new measures: 
(a) \textbf{Interpretability}: 
current metrics can only identify which datasets are of lower indiscriminability while don't present more explanation why it is the case. 
(b)  \textbf{Functionality}:
a dataset with lower discrimination doesn't mean it's useless since the supervision signals provided there can not only help us directly train a system for the specific use case but also provide good supervised transfer for related tasks. Metrics designed in this work focus on the role of discriminating models.

\paragraph{Calls}
Based on observations obtained from this paper, we make the following calls for future research: 
(1) Datasets'  discrimination ability w.r.t top-scoring systems could be included in the dataset schema (such as dataset statement~\cite{bender-friedman-2018-data}), which would allow researchers to gain a saturated understanding of the dataset.
(2) Leaderboard constructors could also report the discriminative ability of the datasets they aim to include. 
(3)  Seldom used datasets are also valuable for model selection, and a more fair dataset searching system should be investigated, for example, relevance- and scientifically meaningful first, instead of other biases, like popularity.

\section*{Acknowledgements}
We would like to thank Graham Neubig and the anonymous reviewers for their valuable comments.
This work was supported by the National Research Foundation of Singapore under its Industry Alignment Fund – Pre-positioning (IAF-PP) Funding Initiative. Any opinions, findings, conclusions, or recommendations expressed in this material are those of the authors and do not reflect the views of the National Research Foundation of Singapore.

\bibliography{custom}
\bibliographystyle{acl_natbib}
\clearpage
\appendix

\section{Statistics of Datasets}
\label{sec:app-statistics}
Tab. \ref{tab:train_dev_test} shows the statistical information of the nine datasets of text classification task used in our work. 
For those datasets without explicit the development set, we randomly selected 
$12.5\%$ samples from the training set as the development set.

\begin{table}[!htp]\centering
\footnotesize
\begin{tabular}{lcccc}\toprule
\textbf{Dataset} & \textbf{Train} & \textbf{Test} & \textbf{Development} \\\midrule
IMDB &25,000 &25,000 &- \\
Yelp &560,000 &38,000 &- \\
QC &5,452 &500 &- \\
DPedia &560,000 &70,000 &- \\
CR &3,594 &400 &- \\
ATIS &4,978 &893 &- \\
SST1 &8,544 &2,210 &1,101 \\
MR &9,596 &1,066 &- \\
ADE &23,516 &- &- \\

\bottomrule
\end{tabular}
\caption{Statistics of datasets.}
\label{tab:train_dev_test}
\end{table}

\section{Parameter Settings for Text Classification Model}
\label{sec:app-param}

In this section, we will introduce the parameter settings of the neural network-based models explored in Section 3.2. The optimizer is AdamW for the four mdoels. The settings of other parameters are shown in Tab. \ref{tab:model}.

\begin{table}[!htp]\centering
\renewcommand\tabcolsep{3pt}
\footnotesize
\begin{tabular}{lrrrrr}\toprule
\textbf{Parameter} &\textbf{BERT} &\textbf{CNN} &\textbf{LSTM} &\textbf{LSTMAtt} \\\midrule
learning rate &2*e-5 &1*e-4 &1*e-3 &1*e-3 \\
batch size &4 &4 &32 &32 \\
word emb &- &Word2vec &GloVe &GloVe \\
word emb size &- &300 &300 &300 \\
hidden size &768 &120 &256 &256 \\
max sent len &512 &- &- &- \\
filter size &- &1,3,5 &- &- \\
\bottomrule
\end{tabular}
\caption{the parameters of four models.}\label{tab:model}
\end{table}

\section{Characterization of Datasets}
\label{sec:app-feat}

\subsection{Inherent Feature}

\noindent\textbf{Label balance ($\phi_{\mathrm{bal}}$):} The label balance metric measures the variance between the ideal and the true label distribution: $\phi_{\text{bal}} = (c_t-c_s)/ c_s$,
where the $c_t$ and $c_s$ are the true and ideal label information entropy \cite{shannon1948mathematical}, respectively.

\subsection{Lexical Feature}

\noindent\textbf{Type-Token Ratio ($\phi_{\mathrm{ttr}}$):} TTR~\citep{richards_1987} is a way to measure the documents lexical richness: $\phi_{\mathrm{ttr}} = n_{\text{type}}/ n_{\text{token}}$, where the $n_{type}$ is the number of unique words, and $n_{token}$ is the number of tokens. We use lexical richness \footnote{\url{https://github.com/LSYS/lexicalrichness}} to calculate the TTR for each sentence and then average them.

\noindent\textbf{Language Mixedness Ratio ($\phi_{\mathrm{lmix}}$):}
The proportion of sentence that contains other languages in the whole dataset. To detect the mixed other languages, we utilize the models proposed by \citet{joulin2016bag} for language identification from fastText~\citep{joulin2016fasttext} which can recognize 176 languages.

\noindent\textbf{Pointwise Mutual Information ($\phi_{\mathrm{pmi}}$):} is a measurement to calculate the correlation between variables.
Specifically, for a word in one class $\phi_{\mathrm{pmi(c,w)}} = \log(\frac{p(c,w)}{p(c)p(w)})$, where $p(c)$ is the proportion of the tokens belonging to label $c$, $p(w)$ is the proportion of the word $w$, and $p(c,w)$ is the proportion of the word $w$ which belongs to class $c$. For every class, all the $\phi_{\mathrm{pmi(c,w)}}$, larger than zero, are added to get the sum, which serve as the dataset's pmi. Finally,$\phi_{\mathrm{pmi}}$ is calculated by dividing the sum by the numbers of pairs(c,w) of the train dataset. We pick up the top-ten words sorted by $\phi_{\mathrm{pmi(c,w)}}$ in all classes, then the ration related to the class-related word($\phi_{\mathrm{r_{pmi}}}$) is calculated by dividing the number of samples who contain the top-ten words by the total samples in the train set.

\subsection{Semantic Feature}

\noindent\textbf{Grammar errors ratio ($\phi_{\mathrm{gerr}}$):} The proportion of words with grammatical errors in the whole dataset. We adopt the detection tool  \footnote{\url{https://github.com/jxmorris12/language_tool_python}} to recognize words with grammatical errors. 
We first compute the grammar errors ratio for each sentence: $n/m$, where the n and m denote the number of words with grammatical errors and the number of the token for a sentence, averaging them.

\noindent\textbf{Flesch Reading Ease ($\phi_{\mathrm{fre}}$):} Flesch Reading Ease \footnote{\url{https://en.wikipedia.org/wiki/Flesch\%E2\%80\%93Kincaid_readability_tests}} calculated by textstat \footnote{\url{https://github.com/shivam5992/textstat}} is a way to describe the simplicity of a reader who can read a text. First, we calculate the $\phi_{\mathrm{fre}}$ for each sample, and then average them as the dataset's feature. 
Then we pick out the samples whose score below 60, then the ration related to the low score samples($\phi_{\mathrm{r_{fre}}}$) is calculated by dividing the number of the picked samples by the total samples in the train set.

\end{document}